\documentclass[runningheads]{llncs}

\usepackage{graphicx}
\usepackage{amsmath,amssymb} 
\usepackage{color}
\usepackage[width=122mm,left=12mm,paperwidth=146mm,height=193mm,top=12mm,paperheight=217mm]{geometry}

\usepackage{booktabs}
\usepackage{xcolor}
\usepackage{adjustbox}
\usepackage{tabularx}
\usepackage{bm}
\usepackage{hyperref}

\usepackage{capt-of}

\definecolor{mint}{rgb}{.63,.79,.95}

\newcommand{\eg}{e.g.}
\newcommand{\ie}{i.e.}
\newcommand{\etal}{et. al.}

\begin{document}
\pagestyle{headings}
\mainmatter
\def\ECCV{12349}  

\title{AddBiomechanics Dataset: \\ Capturing the Physics of Human Motion at Scale} 

\titlerunning{AddBiomechanics Dataset: Capturing the Physics of Human Motion at Scale}

\authorrunning{Werling et. al.}

\author{Keenon Werling$^1$, Janelle Kaneda$^1$, Alan Tan$^1$, Rishi Agarwal$^1$, Six Skov$^1$, Tom Van Wouwe$^1$, Scott Uhlrich$^1$, Nicholas Bianco$^1$, Carmichael Ong$^1$, Antoine Falisse$^1$, Shardul Sapkota$^1$, Aidan Chandra$^1$, Joshua Carter$^2$, Ezio Preatoni$^2$, Benjamin Fregly$^3$, Jennifer Hicks$^1$, Scott Delp$^1$, C. Karen Liu$^1$}
\institute{$^1$Stanford University, $^2$University of Bath, $^3$Rice University}

\maketitle

\begin{abstract}

While reconstructing human poses in 3D from inexpensive sensors has advanced significantly in recent years, quantifying the dynamics of human motion, including the muscle-generated joint torques and external forces, remains a challenge.
Prior attempts to estimate physics from reconstructed human poses have been hampered by a lack of datasets with high-quality pose and force data for a variety of movements.
We present the \emph{AddBiomechanics Dataset 1.0}, which includes physically accurate human dynamics of 273 human subjects, over 70 hours of motion and force plate data, totaling more than 24 million frames.
To construct this dataset, novel analytical methods were required, which are also reported here.
We propose a benchmark for estimating human dynamics from motion using this dataset, and present several baseline results.
The AddBiomechanics Dataset is publicly available at \href{https://addbiomechanics.org/download\_data.html}{addbiomechanics.org/download\_data.html}.

\end{abstract}

\section{Introduction}
\label{sec:intro}

The ability to accurately infer human movement \textit{dynamics} (the motion \textit{and} the physical forces that generated the motion) from video cameras or wearable sensors would transform the fields of computer graphics, robotics, orthopedics, biomechanics, and rehabilitation. With measurement not only of movement but also the internal and external forces driving the movement, researchers can generate more realistic animations of movement \cite{lee2019scalable}, design wearable robotic systems \cite{molinaro2020biological,lee2021biomechanical}, study injury mechanisms \cite{boden2010non,kalkhoven2021training}, and improve treatments for mobility impairments \cite{hunt2011feasibility,killen2020silico}.

\begin{figure}[ht]
\centering
    \includegraphics[width=\textwidth]{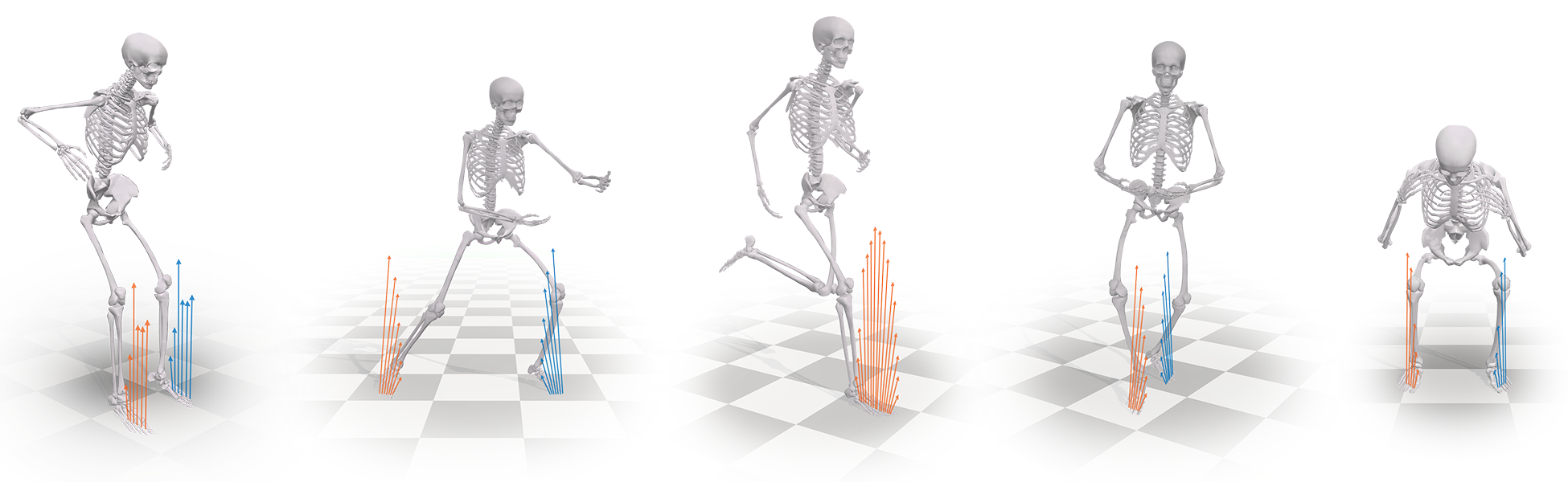}
    \captionof{figure}{Frames from recordings of a drop jump, tai chi, running, ballet, and squatting (left to right) in the AddBiomechanics Dataset. The ground reaction forces are shown with arrows on the right (orange) and left (blue) feet for representative time steps around the shown frame.
    The model includes 22 body segments, 37 degrees of freedom, and specialized joints for the knees, shoulders, and spine, which match the subject's mass, inertial properties, and motion capture recordings while obeying $F=ma$ with respect to the force plate recordings.
    Discrepancies between the reconstruction and the original sensor data are within clinically acceptable tolerances \cite{hicks2015my}.}
    \label{fig:img1}
\end{figure}

To reconstruct human movement dynamics from inexpensive sensors, we need realistic musculoskeletal models of the human body, along with accurate estimates of human pose and external moments and forces. Over the past decade, the research community has made significant progress in both generating realistic musculoskeletal models of the body \cite{rajagopal2016full,holzbaur2005model,vasavada1998influence,silva2004sensitivity} and reconstructing poses from inexpensive sensors such as cameras\cite{goel2023humans,cao2017realtime,toshev2014deeppose,shi2023phasemp,holmquist2023diffpose,li2022mhformer,kocabas2020vibe,lin2021end}, sparse wearable sensors \cite{jiang2023drop}, and generative models \cite{rempe2021humor}. However, the measurement of external forces outside of motion capture labs is unreliable \cite{lambrich2023concurrent,deberardinis2018assessing} and thus far the models that estimate external forces from motion alone are intractable for real time use \cite{falisse2016emg,uhlrich2022opencap} or are not evaluated against measured data \cite{yang2023ppr,guo2023physics,jiang2023drop,shimada2021neural,shimada2020physcap,peng2018deepmimic,vondrak2012dynamical}. Thus their accuracy cannot be compared to the thresholds established in the literature \cite{hicks2015my}.  A key bottleneck is the lack of combined motion and force datasets to train models and assess accuracy of external force estimates against real measurements for a diversity of human subjects and activities.

To address this problem, we present the AddBiomechanics Dataset, the first large scale standardized dataset of lab-based pose and force data for a variety of movements, currently with over 273 participants and 70+ hours of recorded motion.
This dataset can be used to train machine learning methods to reconstruct detailed force information about human movement dynamics (\eg biological torques, ground reaction moments and forces, etc.) from easily measurable quantities (\eg motion capture from a cell phone camera). To provide a fair basis to compare future models that estimate joint torques and external moments and forces acting on the body from motion alone, we propose a set of evaluation metrics, along with a standard train/development/test split. To provide a sense for what is possible with this data, we implement both a simple model and the architecture from previous papers that have attempted data-driven ground reaction moment (GRM) and force (GRF) prediction models \cite{han2023groundlink,mourot2022underpressure}, and report state-of-the-art results on the AddBiomechanics Dataset.

To construct our dataset, we present a key improvement to the AddBiomechanics tool presented by Werling \etal \cite{werling2023addbiomechanics}, which quickly and automatically provides estimates of model scales, motion, and dynamics that meet rigorous standards for analyzing human motion from the biomechanics literature \cite{hicks2015my}. In particular, the original AddBiomechanics tool is unable to process movement trials with a mix of measured and unmeasured external forces, and compute dynamics only when the external forces are available. These trial types are common since, to obtain quantities of diverse raw data recordings at a scale useful for machine learning, researchers must collect during long sessions in labs with motion capture and in-ground force plates, but instrumenting the entire floor of a lab is prohibitively expensive. Long captures of diverse data therefore contain frames where the subject stepped off of the force plates and experienced large unmeasured external ground forces acting on their body. While we still cannot use those frames in the final dataset, we present a method to use the optical motion capture on those ``unobserved forces'' frames to constrain the dynamics reconstruction on frames where all external forces acting on the body \textit{are} measured, and allowing long diverse capture sessions to be optimized together. This improvement allows us to reconstruct accurate dynamics on large, diverse and previously inaccessible raw data sources, making the resulting dataset larger and more diverse.


In summary, we make the following novel contributions:

 \begin{itemize}

\item \textbf{Releasing the largest dataset of human dynamics}, containing personalized musculoskeletal models and corresponding experimental measurements of foot-ground contact forces paired with motion capture data.


\item \textbf{Improving automated human dynamics data processing tools} to enable the construction of a large and diverse dataset, by overcoming a limitation that required that all steps be on force plates. We discard timesteps with steps off of force plates (often present in long continuous sessions), but can still find dynamics on those frames where force plate data is present.

\item \textbf{Proposing a standard benchmark} to evaluate methods that reconstruct human dynamics information from human motion alone. For example, some work \cite{han2023groundlink,mourot2022underpressure} evaluates error only for the vertical ground reaction force, which is a strict lower bound on taking the L2-norm between predicted and measured 3D force vectors.

\end{itemize}

\section{Related Work}

\subsection{Related Datasets}

While there are several valuable, existing datasets including both pose and force-related data, they are currently small-in-scale, of insufficient accuracy, and/or in a format unsuitable for the task of predicting forces (\eg, long captures with only partially observed experimental force data). For example, one of the largest available datasets of human dynamics data with pre-computed full body models is presented in UnderPressure \cite{mourot2022underpressure}, with 5.6 hours of data from 10 subjects. UnderPressure contains joint kinematics, inertial measurement unit (IMU) data and synchronized pressure insole data collected during various types of locomotion. While the largest available, pressure insole data is of insufficient accuracy to create or evaluate 3D dynamic simulations \cite{lambrich2023concurrent,deberardinis2018assessing,camomilla2017methodological}. 

There are several additional datasets, of varying size, with higher-quality force data from expensive in-floor or in-treadmill force plates, but these datasets have heterogeneous processing, incomplete observation data or formatting inconsistencies. Carter \etal \cite{carter2023}  and Camargo \etal \cite{camargo2021comprehensive} each share nearly 20 hours of optical marker motion capture and force data for subjects performing walking, stairs, ramps, running, jumping, and other activities. Most datasets collected with in-ground force plates (\eg, Camargo \etal) have large sections of partial force data as a subject steps onto or off of an in-ground plate, which existing analysis tools cannot handle. Additional human dynamics datasets (most under 2 hours) include video keypoints with synchronized ground force plate data \cite{morris2021predicting}, sparse IMU sets with instrumented treadmills \cite{alcantara2022predicting,lim2019prediction,scheltinga2023estimating}, and traditional optical markers with ground force plates \cite{johnson2018predicting,oh2013prediction,mundt2020prediction,liu2022deep}, with Groundlink \cite{han2023groundlink} representing the widest diversity of dynamics with 19 different motions. We combine these and several additional datasets into the AddBiomechanics Dataset 1.0.

\subsection{Estimating Physical Forces from Motion}
\label{sec:estimating_grf}
In computer vision and graphics, physics (either in the form of a physics engine or constraints to represent "physical intuitions") has been applied by many researchers to improve the quality of generated movements, to combat common artefacts such as foot floating or foot sliding.

One common approach is the use of fast physics-based reinforcement learning methods \cite{peng2018deepmimic,peng2021amp,peng2022ase,ling2020character,bergamin2019drecon,yuan2023physdiff,yi2022physical,yuan2021simpoe,li2021reinforcement,yang2020multi,shimada2020physcap,shimada2021neural}, generally with simplified contact models (\eg, ``rigid box feet'') to estimate joint torques necessary to achieve a measured motion. Another area of focus has been improving the quality of human motion simulations (whether synthesized \textit{de novo} or obtained from a motion capture method) by leveraging ``physical intuitions'' about how humans move in the world (\eg, minimizing foot sliding, disallowing foot-ground penetration, etc.). Approaches to this task have specifically targeted the problem of visual motion quality, whether in isolation \cite{guo2023physics,jiang2023drop,rempe2021humor,rempe2020contact} or tied to a monocular motion capture pipeline \cite{yang2023ppr,li2019estimating,xu2020deep,zell2017joint,brubaker2009estimating,brubaker2007physics}. Both the physics based and physical intuition based methods use “physical plausibility” to describe results and evaluate against motion data alone. Quantifying the accuracy of estimated external forces and joint torques requires measured ground reaction force data with synchronized motion that isn't currently available. The ability of existing methods to reconstruct experimentally valid human dynamics remains an open question.

Combined motion and force datasets have also been used on a small scale to evaluate the models that estimate moments and forces from motion. One approach uses measured motion and force data to reconstruct dynamics with offline trajectory optimization methods and biomechanically accurate models (\ie, using muscle actuators) \cite{haraguchi2023prediction,gartner2022trajectory,uhlrich2022opencap,dembia2020opensim,karatsidis2019musculoskeletal}. Optimization is computationally expensive (\eg, 0.001x real time \cite{karatsidis2019musculoskeletal,uhlrich2022opencap}), can only run on a few seconds of data at a time, has errors on the order of 10\% of body weight compared to measured ground reaction forces \cite{karatsidis2019musculoskeletal,uhlrich2022opencap} and depends on a task-specific movement objective (\eg, minimize energy, maximize speed). A second approach leverages motion and force data to train deep learning models to predict ground reaction forces for specific tasks, such as running \cite{alcantara2022predicting,scheltinga2023estimating}, walking \cite{mundt2020prediction,oh2013prediction,lim2019prediction}, sidestepping \cite{morris2021predicting,johnson2018predicting}, squatting \cite{louis2022learning}, and stair climbing \cite{liu2022deep}. Although comparing the accuracy is challenging due to a lack of standard evaluation metrics, these models perform well for task-specific prediction (\eg, 0.16 BW RMSE \cite{alcantara2022predicting}; \textit{r} = 0.93 for vertical GRF \cite{mundt2020prediction}). There have been few data-driven models trained on multiple activities and evaluated on out-of-distribution movements \cite{louis2022learning,han2023groundlink} and their accompanying datasets are still limited (\eg, less than eight subjects) with average errors exceeding 25\% of body weight. 

In summary, it is currently possible to predict with 10\% error foot-floor contact forces for common activities like walking using offline optimization-based methods \cite{karatsidis2019musculoskeletal}, but no fast, accurate, and generalized models exist.

\section{AddBiomechanics Dataset}
\label{sec:dataset_overview}
\begin{table*}[ht]
\centering
\caption{Breakdown of subject count and hours from the raw data sources.}
\label{tab:dataset-breakdown}
    \begin{tabularx}{0.9\textwidth}{|p{4cm}|X|X|X|}
        \hline 
        \textbf{Source} & \textbf{Subjects} & \textbf{Total Hours} & \textbf{GRF Hours} \\ \hline 
        Lencioni \etal 2019 \cite{lencioni2019human}&  50& 0.52 & 0.37 \\ \hline 
        Carter \etal 2023 \cite{carter2023}&  50& 21.60 & 19.80 \\ \hline 
        Santos \etal 2017 \cite{dos2017data}&  49& 4.90 & 4.90 \\ \hline 
        Camargo \etal 2021 \cite{camargo2021comprehensive}&  22& 19.87 & 10.10 \\ \hline 
        Tan \etal 2023 \cite{tan2023imu}&  17& 4.40 & 4.40 \\ \hline 
        Moore \etal 2015 \cite{moore2015elaborate} &  12& 6.22 & 6.03 \\ \hline 
        Falisse \etal 2016 \cite{falisse2016emg}&  11& 0.49 & 0.10 \\ \hline 
        Hamner \etal 2013 \cite{hamner2013muscle}&  10& 0.02 & 0.02 \\ \hline 
        Van der Zee \etal 2022 \cite{van2022biomechanics}&  10& 5.46 & 5.31 \\ \hline 
        Uhlrich \etal 2023 \cite{uhlrich2022opencap}&  10& 0.24 & 0.05  \\ \hline 
        Tan \etal 2022 \cite{tan2022strike}&  9& 3.73 & 3.73 \\ \hline 
        Wang \etal 2023 \cite{wang2023wearable}&  9& 1.84 & 1.84  \\ \hline
        Han \etal 2023 \cite{han2023groundlink}& 7& 1.70 & 0.52 \\ \hline
        Fregly \etal 2012 \cite{fregly2012grand}& 6& 0.14 & 0.04  \\ \hline
        Li \etal 2021 \cite{li2021well}& 1& 0.34 & 0.34  \\ \hline
        \textbf{Sum}                             & \textbf{273} & \textbf{71.47} & \textbf{57.55} \\ \hline
    \end{tabularx}   
\end{table*}
Version 1.0 of the AddBiomechanics Dataset contains standardized musculoskeletal models as well as position and physics information for over 24 million frames from 70+ hours of motion for 273 participants (Table \ref{tab:dataset-breakdown}). Each of these frames contains optical marker locations and ground reaction moment and force measurements, along with estimated joint kinematics, estimated joint torques, and estimated center of mass kinematics. The dataset includes nine different activity types collected from 15 publicly available raw data sources (Table \ref{tab:dataset-breakdown}), captured in 12 different laboratories. Each raw data source includes experimental optical motion capture and force plate data; refer to the corresponding publications for more details about the experimental data collection. All raw data was run through the same processing pipeline, presented in Section \ref{section:technical_approach}. For the data sources that reported demographic information, the subjects have a mean age of 30.7 ($\pm 15.8$) years (range: $6-84$), and mean BMI of 22.8 ($\pm 3.4$) (range: $11.7-34.4$). The majority of the subjects are male (73\%), and 23\% are female. A portion of the datasets did not contain reported age (9\%) or biological sex (4\%).
        
\begin{figure}[ht]
    \centering
    \includegraphics[width=0.8\linewidth]{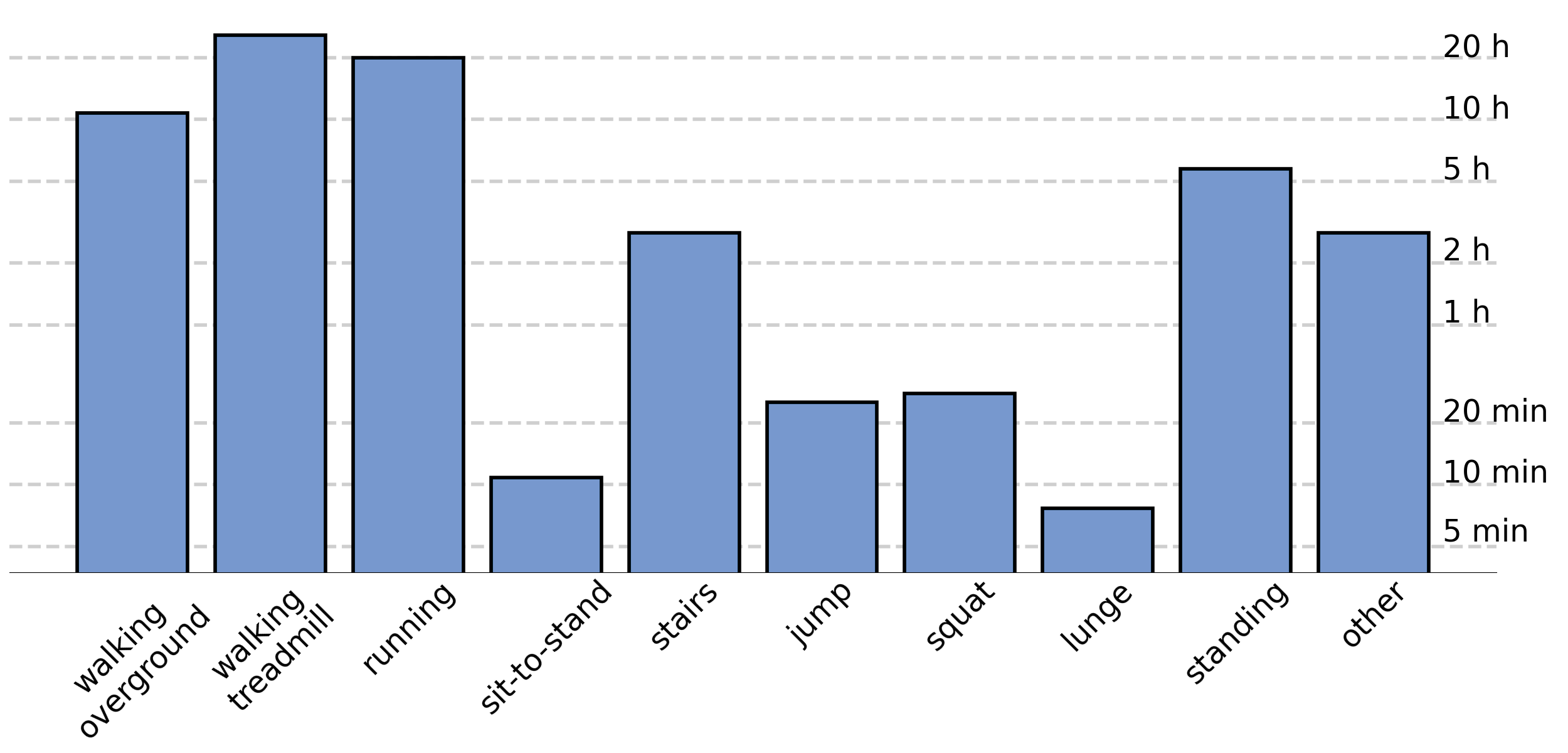}
    \caption{Activity classification. The duration of captures in each activity class is shown on a log scale.} 
    \label{fig:activity_classification}
\end{figure}

We manually classified the activity types in the AddBiomechanics Dataset by visualizing the kinematics of each capture. Labels were generated from the activities presented by each dataset. Sub-labels further classify each activity class; for example, there are sub-labels for overground vs. treadmill walking and running, and for walking up vs. down stairs. Any capture that could not be clearly categorized under the main activity labels is classified as "other." For example, several captures from \cite{moore2015elaborate} are subjects stepping onto a treadmill, and some captures contain calibration motions. Figure \ref{fig:activity_classification} shows the duration of each activity class within the AddBiomechanics Dataset.

\begin{figure}
    \centering
    \includegraphics[width=0.8\linewidth]{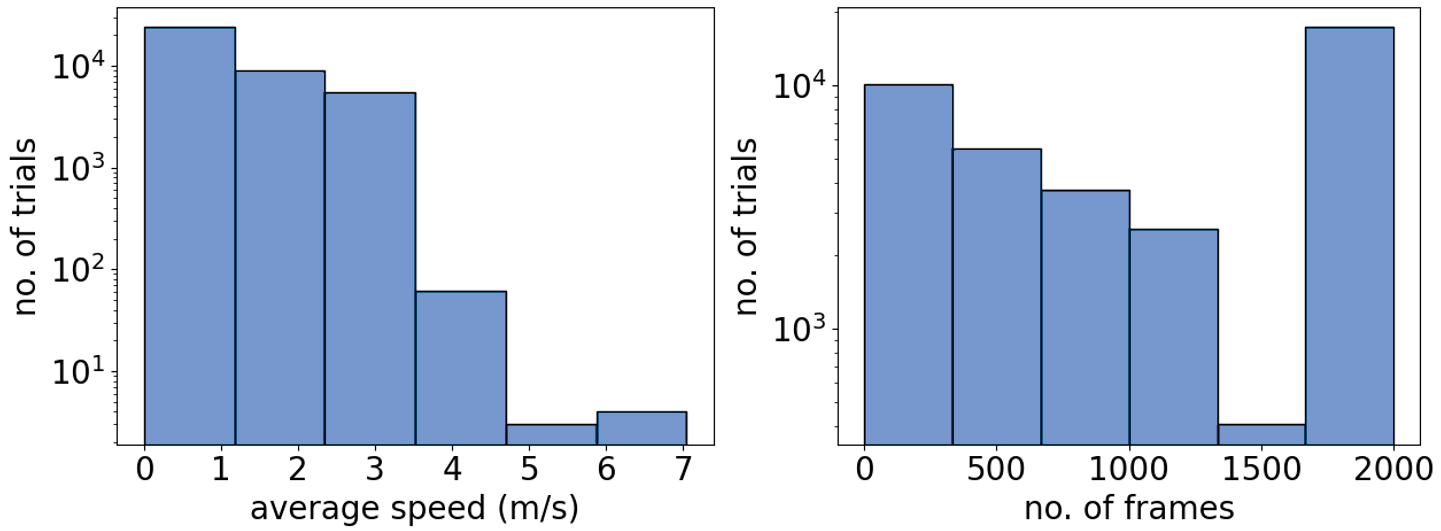}
    \caption{Distributions of speed and capture length in the AddBiomechanics Dataset. The magnitude of speed averaged per trial (\textbf{left}). Trial lengths as number of frames (\textbf{right}). Original trials longer than 2000 frames were split up into multiple trials segments.} 
    \label{fig:trials-histo}
\end{figure}

The AddBiomechanics Dataset contains a variety of speeds and individual trial lengths (Figure \ref{fig:trials-histo}). Averaged over each trial, the majority of absolute speeds fall between about 0.0 and 5.0 m/s. Trials were split into segments of at most 2,000 frames as a result of computational limits in the processing pipeline. The majority of remaining trials contain less than 1,000 frames. Overall, 53.4\% of the GRF in the dataset is single support, followed by 21.2\% for double support and 25.4\% for flight phase. 

\subsection{Personalized Subject Models}

We provide individual subject skeletons as scaled and mass-optimized versions of the Rajagopal Full Body Model \cite{rajagopal2016full}, which is the standard in human biomechanics for maximum biomechanical accuracy in modelling the physics of the body. Recent work by Keller \etal \cite{keller_skel} means the Rajagopal model can also be transformed to a variant of the SMPL Model, ensuring that the AddBiomechanics Dataset can be used with existing SMPL-based datasets (\eg \cite{AMASS:ICCV:2019,ionescu2013human3}) and methods (\eg \cite{rempe2021humor,goel2023humans,yang2023ppr,guo2023physics,jiang2023drop,shimada2021neural,shimada2020physcap,peng2018deepmimic,vondrak2012dynamical}). 

\subsection{Evaluating Dataset Quality}

\begin{figure}[ht]
\centering
    \includegraphics[width=\textwidth]{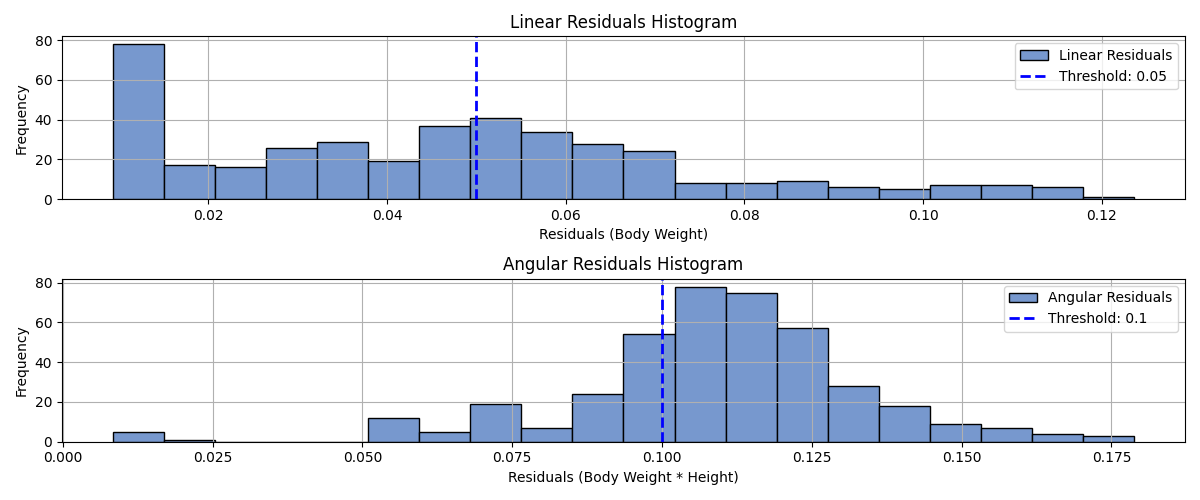}
    \captionof{figure}{Evaluating dataset quality. See Table \ref{tab:hicks-threshold} for descriptions of quantities.}
    \label{fig:residuals_histo}
\end{figure}

It is best practice in biomechanics to evaluate dynamic reconstructions of human movement from raw sensor data (marker locations, force plate data) using the discrepancy between the raw measurements and what the dynamic model implies the measurements should be \cite{hicks2015my}. These standard measurements are available in Table \ref{tab:hicks-threshold}. There are thresholds for clinical grade dynamics suggested by Hicks \etal \cite{hicks2015my}, also reported in the table. Many human modelling experts do not reach this bar in practice \cite{werling2023addbiomechanics}, so passing these thresholds at scale is a significant achievement. 12.1 hours (21.2\%) of the data is classified as clinical-grade human dynamics data by this measure, an unprecedentedly large dataset to reach that quality bar. For overall distributions of residuals, see Figure \ref{fig:residuals_histo}.



\begin{table*}[ht]
    \centering
    \begin{tabular}{l|c|c}
        Measurement & Mean $\pm$ Std-dev & Hicks Threshold for Clinical Use \cite{hicks2015my} \\
        \hline
        Marker Error & $2.17\pm2.44$ (cm) & N/A \\
        \hline
        Linear Residual & $0.046\pm0.027$ (BW) & 0.05 BW \\
        \hline
        Angular Residual & $0.11\pm0.02$ (BW*h) & 0.1 BW*h \\
    \end{tabular}
    \vspace{1.0em}
\caption{Evaluating dataset quality. ``Marker Error'' is measured in centimeters RMS to evaluate the quality of pose reconstruction. ``Linear Residual'' is force discrepancy between model and raw data, normalized across subjects as body-weights (BW) RMS. ``Angular Residual'' is torque dynamic discrepancy at the model's root, normalized across subjects as body-weights * height (BW*h) RMS. The Hicks Thresholds \cite{hicks2015my} for clinical model accuracy are presented with each metric.}
    \label{tab:hicks-threshold}
\end{table*}

\section{Raw Data Processing and Aggregation}
\label{section:technical_approach}

Raw data sources contain optical marker traces, and time synced force plate data. It requires a substantial optimization problem to reconstruct human dynamics. The backbone of our approach is to manually review our raw data sources frame-by-frame to determine frames where steps off of force plates occurred, and then run the annotated raw data through the pipeline presented in Werling \etal \cite{werling2023addbiomechanics}, with the modifications described below to handle steps off of force plates.

\begin{figure}[ht]
\centering
    \includegraphics[width=\textwidth]{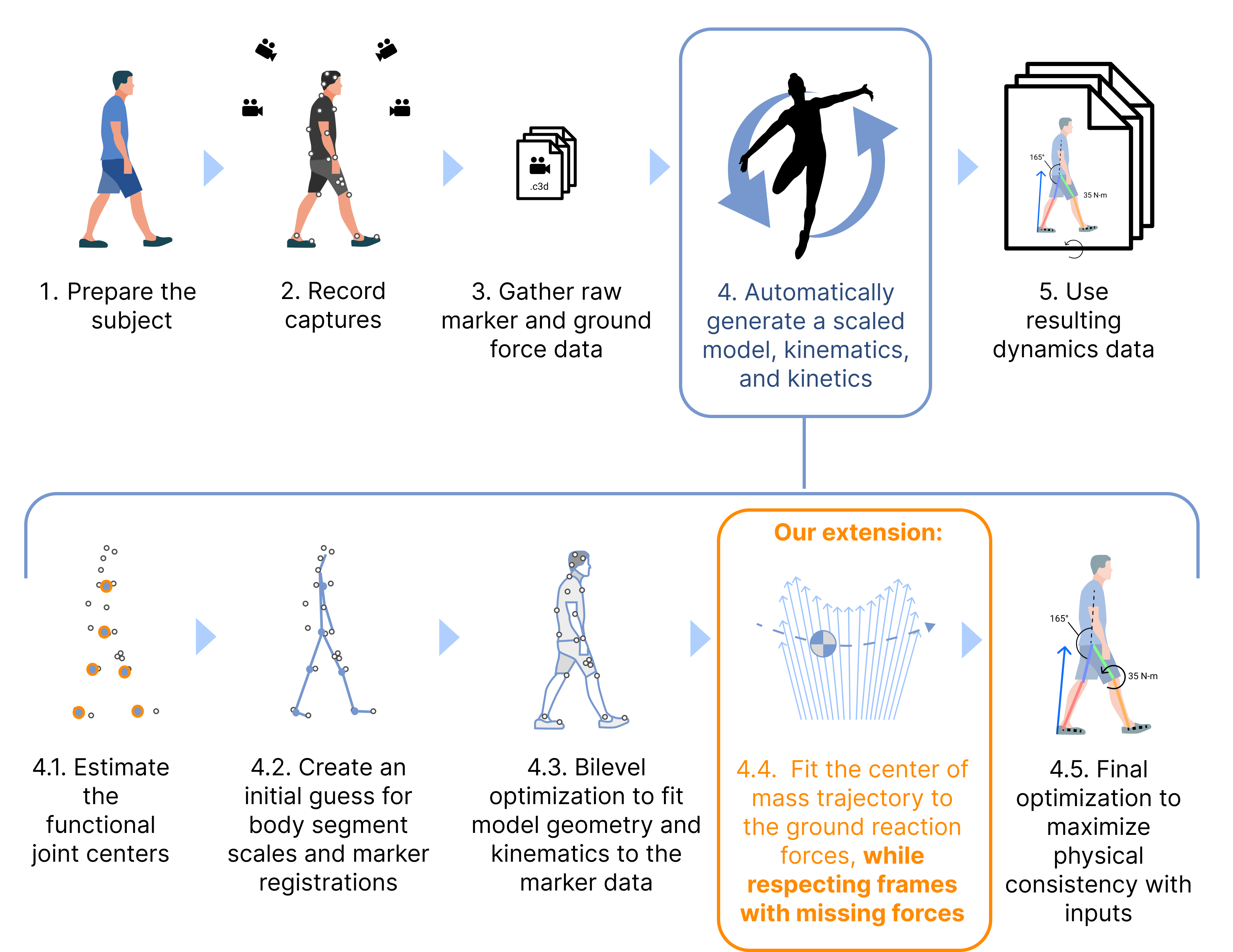}
    \captionof{figure}{Adapted with permission from Werling \etal \cite{werling2023addbiomechanics}, this figure shows the processing pipeline we used to turn raw sensor data into accurate human dynamics, and shows where the method described below swaps into the original pipeline from \cite{werling2023addbiomechanics}}
    \label{fig:addb_pipeline}
\end{figure}

In collecting large amounts of motion data with overground force plates, it is common for subjects to step on uninstrumeted ground and have unmeasured forces acting on them several times during a single capture. These frames of data are not useful for learning about human dynamics, so should be thrown away, but that must be done after we have solved for dynamics on the remaining frames. If we were to simply delete the frames, and solve for skeleton scaling and motion dynamics separately on each range of frames with continuously observed forces, we could get quite unrealistic motion, since there are no useful constraints on the start and end conditions of each short group of frames with observed forces, and so the optimizer can cheat with unrealistic initial and terminal velocities. Optimizing for dynamics with the motion from frames with unobserved force data helps prevent that from happening.

In order to construct this dataset, we needed to be able to solve for accurate human dynamics while ignoring physical consistency on frames with steps off of force plates, and yet not allowing any discontinuous jumps in human kinematics when they step off of or back onto force plates. This section describes the relevant background, and our algorithms to handle this problem.

These methods are enhancements to the methods originally presented in Werling \etal \cite{werling2023addbiomechanics} These algorithms are already merged into the upstream Werling \etal \cite{werling2023addbiomechanics} tool, and are available in the open source codebase.

\subsection{Processing Raw Sensor Data}
\label{addb_bg}

Biomechanical motion capture labs output optical marker traces over time, and time-synced ground reaction force measurements (which include force, locations, and torque for each contact) from force plates in the ground or in treadmills. It is possible to run inverse kinematics on the locations of the optical markers over time to produce an accurate reconstruction of motion by picking joint angles $\bm{q}$ and bone scales $\bm{s}$ to minimize the distance between virtual markers $f(\bm{q}, \bm{s})$ and observed markers $\bm{o}$.

\begin{eqnarray}
\min_{\bm{q}} ||f(\bm{q}, \bm{s}) - \bm{o}||
\end{eqnarray}

It is also possible, given a skeleton model, joint accelerations and observations of external forces $\bm{f}_t$ from the force plates, to run standard inverse dynamics to compute joint torques.
\begin{eqnarray}
\label{eq:inverse_dynamics}
\bm{\tau}_t = \bm{M}(\ddot{\bm{q}}_t) - \bm{C}_t - \bm{J}^T\bm{f}_t
\end{eqnarray}

We adopt standard notation that $\bm{M}$ is the mass matrix, $\bm{C}_t$ is the vector of joint torques from coriolis forces and gravity, $\bm{J}$ is the jacobian of the contacts, and $\bm{f}_t$ is the contact forces.

Here $\ddot{\bm{q}}_t$ is estimated by central differencing the joint angles, with the equation given below.
\begin{eqnarray}
\label{eq:central_difference}
\Tilde{\ddot{\bm{q}}}_t = \frac{\Tilde{\bm{q}}_{t-1} - 2\Tilde{\bm{q}}_t + \Tilde{\bm{q}}_{t+1}}{\Delta t^2}
\end{eqnarray}
It is worth noting that noise in motion capture systems that estimate our $\bm{q}_t$ values tends to have very high frequency content, because each frame's errors are approximately independently and identically distributed at 100fps. This leads to a signal to noise ratio that is 4 orders of magnitude worse after finite differencing. We use a Butterworth low pass filter at 30Hz to process the finite differenced estimates, which does a good job attenuating the noise in our estimates of $\ddot{\bm{q}}_t$ but makes them only trustworthy in their lower frequency components.

Returning to Equation \ref{eq:inverse_dynamics}, the problem now arises that in order to be physically possible, whichever degrees of freedom correspond to the free translation and rotation of the skeleton in the world (generally the first 6, 3 for translation and 3 for rotation) in $\bm{\tau}$ must be 0. This constraint (\ie, $\bm{\tau}^{[0:6]} = 0$) reflects the fact that all external forces are applied by the force plate, and not by a giant invisible robot arm at the root segment. However, there is no guarantee in the standard inverse dynamics equations that the first 6 entries of $\bm{\tau}_t$ will be 0. To the extent they are not zero, they have typically been called ``residual forces'' in the literature.

The goal in producing realistic digital twins of subjects is to find subject bone scales $\bm{s}$, positions over time $\bm{q}_t$, and bone masses and inertial properties $\bm{m}$ such that the residual forces are minimized when we solve inverse dynamics. Leaving out some less important variables for clarity, Werling \etal \cite{werling2023addbiomechanics} optimize the following non-convex problem to solve for $\bm{s}$, $\bm{q}_t$, and $\bm{m}$:

\begin{eqnarray}
\label{eq:kitchen_sink}
\min_{\bm{q}_t, \bm{s}, \bm{m}} \sum_t \underbrace{||f(\bm{q}_t, \bm{s}) - \bm{o}_t||}_{\text{Inverse Kinematics}} + \underbrace{||\bm{\tau}_t^{[0:6]}||}_{\text{Inverse Dynamics}}
\end{eqnarray}

The challenge with this approach is that it requires a good initial guess for the values of $\bm{q}_t$, $\bm{s}$, $\bm{m}$ because the optimization problem is non-convex.

In the original method, these initial guesses for $\bm{q}_t$ and $\bm{s}$ are constructed through a series of linear and convex optimization problems with marker data. The $\bm{q}_t$ initialization is then adjusted so the center of mass trajectory implied by $\bm{q}_t$ is consistent with the trajectory dictated by the ground reaction forces, defined by the differential equation:
\begin{eqnarray}
\label{eq:com_diffeq}
\ddot{\bm{z}} = \frac{\bm{f}}{m} - \bm{g}
\end{eqnarray}
where $\ddot{\bm{z}} \in \mathbb{R}^3$ is center of mass acceleration, $\bm{f}$ is the ground reaction force vector, $m$ is the system mass, and $\bm{g}$ is gravitational acceleration.

\subsection{Fitting Continuous Dynamics While Ignoring Frames with Missing Force Data}
\label{addb_extensions}

When all the ground reaction forces are known from experimental data, then the entire trajectory $\bm{z}_t$ is linearly determined by the inverse of mass $\mu$, $\bm{z}_1$ and $\dot{\bm{z}}_1$. However, given a set of frames where the external forces are no longer observed, $\bm{U} = \{\bm{u}_1 \in \mathbb{N}, \bm{u}_2 \in \mathbb{N}, \ldots\}$, then this simple linear relationship ceases to hold. The center of mass trajectory $\bm{z}_t$ is now also dependent on the total external forces on those frames, divided by the unknown subject mass, then integrated over time, which makes the relationship between $\mu$, $\bm{z}_1$, $\dot{\bm{z}}_1$, and all $\bm{f}_{\bm{u}_i}$ for $\bm{u}_i \in \bm{U}$ quadratic.

To restore linearity in the presence of unknown forces, we improve the original Werling \etal method by explicitly solving for the center of mass acceleration on each unobserved frame: $\ddot{z}_{\bm{u}_i}$ for all $\bm{u}_i \in \bm{U}$. We define an extended vector $\bm{\zeta}$ that contains our unknowns:

\begin{eqnarray}
\label{eq:path_vector}
\bm{\zeta} = \begin{bmatrix}
    \bm{z}_1 &
    \dot{\bm{z}}_1 &
    \mu &
    \ddot{z}_{\bm{u}_1} &
    \ddot{z}_{\bm{u}_2} &
    \hdots
\end{bmatrix} \in \mathbb{R}^{7 + 3|\bm{U}|}
\end{eqnarray}

We define a linear system with matrix $\bm{A} \in \mathbb{R}^{3(T + |U|) \times 7 + 3|U|}$ and offset $\bm{b} \in \mathbb{R}^{3(T + |U|)}$ that maps the vector $\bm{\zeta}$ onto $\bm{\mathcal{Z}} \in \mathbb{R}^{3T}$, a vector of concatenated center of mass position vectors over time, and $\ddot{\bm{\mathcal{Z}}_{\bm{U}}}$ the concatenated $\ddot{z}_{\bm{u}_i}$'s, scaled by a regularization term $\alpha$:

\begin{eqnarray}
\label{eq:linear_map}
\bm{A}\bm{\zeta} + \bm{b} = \begin{bmatrix}\bm{\mathcal{Z}} & \alpha \ddot{\bm{\mathcal{Z}}_{\bm{U}}}\end{bmatrix} = \begin{bmatrix}
    \bm{z}_1 &
    \bm{z}_2 &
    \hdots &
    \bm{z}_T &
    \alpha \ddot{z}_{\bm{u}_1} &
    \hdots &
    \alpha \ddot{z}_{\bm{u}_{|\bm{U}|}}
\end{bmatrix} \in \mathbb{R}^{3(T + |\bm{U}|)}
\end{eqnarray}

We can construct $\bm{A}$ and $\bm{b}$ using a semi-explicit Euler integration scheme of the center of mass trajectory to relate the unknowns $\bm{\zeta}$ to the center of mass positions, $\bm{\mathcal{Z}}$. We proceed in blocks:

\begin{eqnarray}
\label{eq:A_matrix}
 \bm{A} = \begin{bmatrix}
    \bm{A}_{\bm{\mathcal{Z}}, 3} && \bm{A}_{\bm{\mathcal{Z}}, \ddot{\bm{\mathcal{Z}}}_{\bm{U}}} \\
    0 && \alpha I
\end{bmatrix}
\end{eqnarray}

Here, $\bm{A}_{\bm{\mathcal{Z}}, 3}$ represents the contributions from $\mu$ ,$\bm{z}_1$ and $\dot{\bm{z}}_1$ to the trajectory, similar to the original adjustment in Werling \etal\cite{werling2023addbiomechanics}. Then, $\bm{A}_{\bm{\mathcal{Z}}, \ddot{\bm{\mathcal{Z}}}_{\bm{U}}}$ is the matrix block relating our acceleration on time steps where there are unobserved external forces is the simple time integration of the change in velocity generated by our $\ddot{\bm{z}}_{\bm{u}_i}$ for each unobserved time step $i$, given as follows:

\begin{eqnarray}
\label{eq:A_Z_ddot_Z}
 \bm{A}_{\bm{\mathcal{Z}}, \ddot{\bm{\mathcal{Z}}}_{\bm{U}}}^{[:, i]} = \begin{bmatrix}
    0 &
    \hdots &
    0 &
    \Delta_t^2 I &
    2\Delta_t^2 I &
    \hdots &
    (T - \bm{u}_i) \Delta_t^2 I
\end{bmatrix}
\end{eqnarray}

The vector $\bm{b}$ is as expressed in the original method, now concatenated with a 0 vector in $\mathcal{R}^{3|\bm{U}|}$, so that $\bm{b} \in \mathcal{R}^{3T + 3|\bm{U}|}$.

Using Equation \ref{eq:linear_map}, we apply the pseudo inverse of $\bm{A}$ to obtain a least-squares best fit of the initial conditions and mass of the system, $\hat{\bm{\zeta}}=\bm{A}^\dagger(\bm{\mathcal{Z}}-\bm{b})$, along with all the $\alpha$ regularized center of mass accelerations on time steps with unobserved external forces $\bm{u}_i \in \bm{U}$ and can continue with the original adjustment process \cite{werling2023addbiomechanics}. This adjusted initialization then serves as an initialization for the final problem in Equation \ref{eq:kitchen_sink}, which we adapt slightly as Equation \ref{eq:kitchen_sink_2} by only including the inverse dynamics loss on frames with observed forces.

\begin{eqnarray}
\label{eq:kitchen_sink_2}
\min_{\bm{q}_t, \bm{s}, \bm{m}} \sum_t \underbrace{||f(\bm{q}_t, \bm{s}) - \bm{o}_t||}_{\text{Inverse Kinematics}} + \underbrace{\mathbb{I}(t \notin \bm{U}) ||\bm{\tau}_t^{[0:6]}||}_{\text{Observed Inverse Dynamics}}
\end{eqnarray}

This new algorithm vastly expands the available data with which to construct the dataset by enabling the platform released by Werling \etal \cite{werling2023addbiomechanics} to process motion data with observed external forces on only some time steps.

\subsection{Ablation Study}
\label{sec:ablation}

To evaluate the impact of our improvements to initialization, we ran an ablation study. We started with a long continuous 50 second trial with fully observed ground reaction forces, so ground truth was available. We then artificially hid the ground reaction force data for every 3rd step on the right foot, to simulate stepping off of the force plates. We compared our method (which could see all force data) as the ``oracle'', our proposed method with hidden data ``our method'', and finally the ablated method of running the ``classic'' AddBiomechanics \cite{werling2023addbiomechanics} Stage 4.4 (Fig 4) independently on each sub-segment with observed ground reaction forces ``piecewise baseline''.

\begin{table}[t]
    \centering
    \begin{tabular}{|c|c|c|}
        \hline
         Method & Linear Residual & Marker RMS \\
         \hline
         Oracle & 12.9 N $\pm$ 1.6N & 1.5cm $\pm$ 0.1cm \\
         Our Method & 18.0 N $\pm$ 2.5N & 2.4cm $\pm$ 0.5cm \\
         Piecewise Baseline & 3630N $\pm$ 3220N & 7.7cm $\pm$ 6.2cm \\
        \hline
    \end{tabular}
    \vspace{0.5em}
    \caption{The results of an ablation study (see Section \ref{sec:ablation}) to evaluate the importance of our proposed improvements for initializing center of mass trajectory when there is hidden ground reaction force data. Higher numbers are worse, and indicate a less accurate fit of the data.}
    \label{tab:my_label}
\end{table}

\begin{figure}[t]
  \centering
  \includegraphics[width=0.97\linewidth]{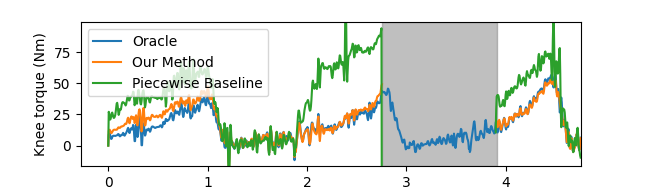}
   \caption{Sample recovered right knee joint torque profiles in ablation studies (see Section \ref{sec:ablation}) of our proposed method. The ``oracle'' blue line is as close as we can get to ground truth, with full access to hidden ground reaction force measurements.}
   \label{fig:ablation_torque}
\end{figure}

The results in Table 1 make it clear that enforcing kinematic smoothness in the initialization is crucial. The non-convex optimizer (Stage 4.5 in Fig 4) does an extremely poor job with the non-smooth initialization it receives from stitching together a number of independent initializations using the ``classic'' AddBiomechanics. The accelerations implied by those large discontinuous position jumps lead to impossibly enormous implied forces. In the process of removing those peaks the optimizer often destroys much of the information content of the initialization, and then is trapped in a poor local optimum and unable to recover. See Figure \ref{fig:ablation_torque} for example torque profiles over time to get a qualitative sense for results.

\section{Benchmark for Estimating Dynamics from Motion}

The key evaluation task of interest enabled by the AddBiomechanics Dataset is estimating physical forces from an observed movement. In this section, we describe in more detail the evaluation task, provide standard metrics relevant to Computer Science and Biomechanics to assess the accuracy of new models, and provide some initial baseline models and their performance on the evaluation metrics.

\subsection{Task Description}

The task assumes an ``off the shelf'' motion capture model (\eg \cite{goel2023humans,jiang2023drop,cao2017realtime,toshev2014deeppose}) produces a noisy time series of joint angles over time, $\Tilde{\bm{q}_t}$ for $t \in [0, T]$. We also assume that a scaled human body model is available for the subject - modern motion capture algorithms provide this, \eg \cite{goel2023humans}. In processing the AddBiomechanics Dataset and in the baselines discussed below, we use the model of Rajagopal \etal \cite{rajagopal2016full}; future models to solve the task could employ this or other skeletal models. 

To construct a full physical simulation of our subject, we now need to solve inverse dynamics (Equation \ref{eq:inverse_dynamics}) using estimated joint accelerations $\ddot{\bm{q}}_t$, and either contact forces and moments acting at the origin of the foot $\bm{f}_t$ or joint torques $\bm{\tau}_t$. The simplest way to achieve this is using central differencing (Equation \ref{eq:central_difference}) and filtering on our input pose estimates $\Tilde{\bm{q}_t}$ to estimate $\ddot{\bm{q}}_t$ and using a model to estimate contact forces and moments $\bm{f}_t$, then solving Equation \ref{eq:inverse_dynamics} for compute joint torques $\bm{\tau}_t$. However, this is only one of many possible approaches to estimating these quantities.

To summarize, the inputs are $\Tilde{\bm{q}_t}$ for $t \in [0, T]$ (a time series of joint angles) and a body model for the subject with correctly scaled bones. And outputs are all three of the following which are consistent with Equation \ref{eq:inverse_dynamics}:
\begin{enumerate}
    \item $\ddot{\bm{q}}_t$ for $t \in [0, T]$: A time series of joint accelerations.
    \item $\bm{f}_t[:3]$ for $t \in [0, T]$: A time series of external moments acting at the origin of the foot.
    \item $\bm{f}_t[3:]$ for $t \in [0, T]$: A time series of external forces.
    \item $\bm{\tau}_t$ for $t \in [0, T]$: A time series of joint torques.
\end{enumerate}

\subsection{Proposed Metrics}

Previous work has used a diversity metrics, measuring different aspects of the problem (\eg r-value for vertical GRF \cite{falisse2016emg}, RMSE for vertical GRF \cite{mourot2022underpressure,han2023groundlink}). We propose a superset of previous metrics. We also include ground reaction moment, which is often neglected in the literature despite contributing significantly to joint torque estimates.

Two clear metrics to evaluate the accuracy of models are the average L2-norm error for our ground reaction force and moment estimate $\hat{\bm{f}}_t$ and average element-wise absolute error for the joint torque estimate $\hat{\bm{\tau}}_t$. Because estimates for $\ddot{\bm{q}}_t$ are not experimentally observed, it does not make sense to include RMSE on $\ddot{\bm{q}}_t$ as a metric. However, we can partially evaluate the quality of $\ddot{\bm{q}}_t$ by checking the RMSE of  $\hat{\ddot{\bm{z}}}_t$, the center of mass acceleration estimated by $\ddot{\bm{q}}_t$. 

To summarize, we propose four evaluation metrics, corresponding to the four outputs of the models, where ground truth values from the dataset are denoted $\ddot{\bm{z}}_t$, $\bm{f}_t$, and $\bm{\tau}_t$ (see \ref{sec:dataset_overview}), and $T$ is the number of time steps and $N$ is the number of degrees of freedom in the skeleton:

\begin{enumerate}
    \item Center of Mass Acceleration: $\sum_t ||\ddot{\bm{z}}_t - \hat{\ddot{\bm{z}}}_t||_2 / T$
    \item Ground Reaction Moment (GRM at foot): $\sum_t ||\bm{f}_t[:3] - \hat{\bm{f}}_t[:3]||_2 / T$
    \item Ground Reaction Force (GRF): $\sum_t ||\bm{f}_t[3:] - \hat{\bm{f}}_t[3:]||_2 / T$
    \item Joint Torques: $\sum_t ||\bm{\tau}_t - \hat{\bm{\tau}}_t||_1 / NT$
\end{enumerate}

\subsection{Preliminary Baseline Experiments}

To set a baseline for future work to improve upon, we tried several trivial or previously reported models on the data. All of these models predict $\bm{f}$ given $\Tilde{\bm{q}}$, and then use Equation \ref{eq:inverse_dynamics} to compute $\bm{\tau}_t$. We evaluate these baselines on the held-out test set (Table \ref{tab:baseline-results}). Brief details of each baseline method are reported below. 

\begin{table*}[ht]
    \centering
    \begin{tabular}{l|c|c|c|c}
        \multicolumn{1}{c|}{Method} & CoM Acc. & Joint Torque & GRM (at foot) & GRF \\
        \hline
        Analytical & 4.47 $m/s^2$ & 2.77 $Nm/kg$ & 6.15 $Nm/kg$ & 2.79 $N/kg$  \\
        MLP Baseline & 1.41 $m/s^2$ & \textbf{1.33 Nm/kg} & \textbf{0.03 Nm/kg} & 1.29 $N/kg$ \\ 
        GroundLinkNet on AddB & \textbf{1.40 m/s$^2$} & 1.34 $Nm/kg$ & 0.07 $Nm/kg$ & \textbf{1.17 N/kg} \\
        \hline\hline
        UnderPressure \cite{mourot2022underpressure} & N/A & N/A & N/A & $>$ 5.74 $N/kg$ \\
        GroundLinkNet \cite{han2023groundlink} & N/A & N/A & N/A & $>$ 3.06 $N/kg$ \\
        Trajectory Optimization \cite{karatsidis2019musculoskeletal} & N/A & N/A & N/A & $\sim 1$ $N/kg$ \\
    \end{tabular}
    \vspace{0.5em}
\caption{Above the double-line is accuracy of simple baseline methods and the model from \cite{han2023groundlink} retrained on the AddBiomechanics dataset, with errors presented as means over the test set. Below the double-line are errors self-reported by previous work. These are not perfectly comparable to our baseline evaluations (or each other), as they each evaluate on their own different test sets, and use their own metrics. Where precise conversion to $N/kg$ errors is not possible (for example, authors only report error on the y-component, or only report $r$-value), approximate numbers or lower bounds are included to give the reader a general sense of approximate relative performance.}
    \label{tab:baseline-results}
\end{table*}

\textbf{Analytical Baseline:} To evaluate performance without machine learning or optimization, we construct a simple baseline method. We estimate $\ddot{\bm{q}}_t$ using Equation \ref{eq:central_difference}. Then we estimate center of mass acceleration $\ddot{\bm{z}}_t$ using $\ddot{\bm{q}}_t$. The total external force acting on the body is estimated using $F = ma$. Then we divide that force up between the feet to create ground force $\bm{f}_t[3:]$, splitting the force evenly during double support. We set ground moment $\bm{f}_t[:3] = 0$.

\textbf{Multi-Layer Perceptron Baseline:} We train a simple two-layer MLP with a hidden state size of 512 to predict $\bm{f}_t$ from $\bm{q}_t$. We use as input features a window of recent $\bm{q}_t$, our estimates of $\ddot{\bm{q}}_t$ from Equation \ref{eq:central_difference}, and joint center locations in the pelvis reference frame. We then use our prediction for $\bm{f}_t$ to compute $\ddot{\bm{z}}_t = \sum \bm{f}_t / m$, and adjust the root-translation coordinates of $\ddot{\bm{q}}_t$ to reflect our new $\ddot{\bm{z}}_t$.

\textbf{Previous Work - GroundLinkNet:} Two previous papers \cite{han2023groundlink,mourot2022underpressure} have tackled the problem of predicting $\bm{f}_t$ (though not $\ddot{\bm{q}}_t$ or $\bm{\tau}_t$) using a similar convolution-based architecture, which we refer to as ``GroundLinkNet'' (where the two models differ we use \cite{han2023groundlink}). For fair comparison, we perform the same procedure as our MLP baseline to predict all three of $\bm{f}_t$, $\ddot{\bm{q}}_t$, and $\bm{\tau}_t$ from just $\bm{f}_t$: the inputs, outputs, and post-processing equations are all identical, the only difference is the model architecture is more complex and larger than the MLP. The model is trained and tested on the same data.

\section{Conclusion and Future Work}

We present a large standardized dataset of high quality human dynamics. This dataset is currently biased towards walking and running, and we plan to release a follow-up dataset with a wider diversity of tasks.

We lay the groundwork for using this data to train models that reconstruct accurate human dynamics information from simple motion capture. We are excited to see the integration of continued progress in inexpensive motion capture systems with methods to infer physics from motion.

We also anticipate other exciting uses of the dataset. It could be possible to use estimates of foot-ground contact and forces derived from this data to remove artefacts from motion capture, like foot sliding. Another exciting application area for the data is ``real to sim'', developing more accurate human body simulators by learning better foot-ground contact colliders from data. This data could also be a useful additional input to motion models, and could have applications in physical motion auto-encoders.

\clearpage

\bibliographystyle{splncs}
\bibliography{egbib}
\end{document}


\section{Supplementary Materials}

\subsection{Noisy Derivatives of Motion}
\label{sec:derivatives}
The standard approach to get second derivatives of motion is to use central differencing on our noisy input pose estimates $\Tilde{\bm{q}}_t$'s:

\begin{eqnarray}
\label{eq:central_difference}
\Tilde{\ddot{\bm{q}}}_t = \frac{\Tilde{\bm{q}}_{t-1} - 2\Tilde{\bm{q}}_t + \Tilde{\bm{q}}_{t+1}}{\Delta t^2}
\end{eqnarray}

The problem with simply relying on central differencing to get $\ddot{\bm{q}}_t$ is that the signal-to-noise ratio gets approximately 4 orders of magnitude worse when we have data at 100fps. To see this, consider that our estimates for center of mass over time have some sort of constant noise offset $\Tilde{\bm{q}}_t = \bm{q}_t + \epsilon_t$, where $\epsilon_t$ is some arbitrary noise vector. We can rearrange Equation \ref{eq:central_difference} to make the noise terms explicit:

\begin{eqnarray}
\Tilde{\ddot{\bm{q}}}_t = \frac{\bm{q}_{t-1} - 2\bm{q}_t + \bm{q}_{t+1}}{\Delta t^2} + \underbrace{\frac{1}{\Delta t^2}}_{\text{Large!}}(\epsilon_{t-1} - 2\epsilon_t + \epsilon_{t+1})
\end{eqnarray}

The multiplication of the noise terms by $(1/\Delta t^2) = 10,000$ when $\Delta t = 0.01$ at 100fps ends up making our estimates of $\Tilde{\ddot{q}}_t$ very poor, often estimating joint accelerations that would physically tear a human apart. Thus, smoothing is essential to get realistic values.

We used the improved AddBiomechanics algorithm to generate a musculoskeletal model scaled to each inputted subject, along with a corresponding simulation of the kinematics and dynamics optimized to match observed optical motion capture and force plate data.  

\subsection{The Importance of Consistent Filtering}
\label{sec:consisten_filtering}
\begin{figure}
    \centering
    \includegraphics[width=0.75\linewidth]{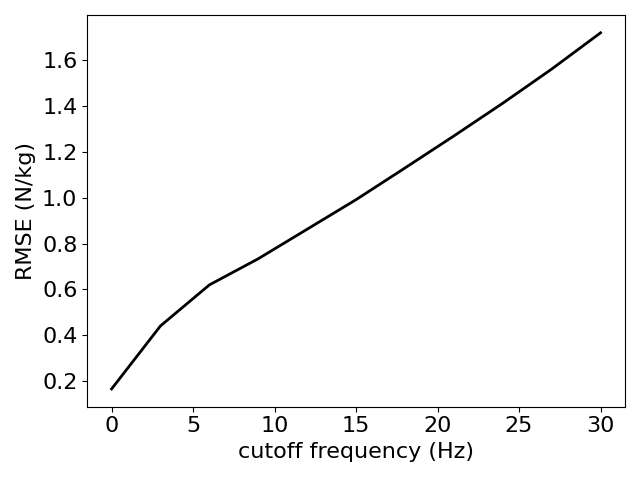}
    \caption{It is important to evaluate the accuracy of models on the ground reaction force (GRF) prediction task with a consistent amount of smoothing in the raw signal data. Previous work has evaluated on custom datasets with inconsistent amounts of smoothing, but this plot shows that as smoothing becomes more aggressive (\textit{i.e.}, cutoff frequency decreases), the GRF prediction error decreases, even with a simple baseline estimation method (using the kinematically-derived center of mass acceleration as in $F = ma$, in this case). The trend of decreasing cutoff frequency with decreasing error highlights the need for consistent filtering for fairly comparing models across datasets. Data from the test set is shown.} 
    \label{fig:err-v-freq}
\end{figure}

\begin{figure}
    \centering
    \includegraphics[width=0.75\linewidth]{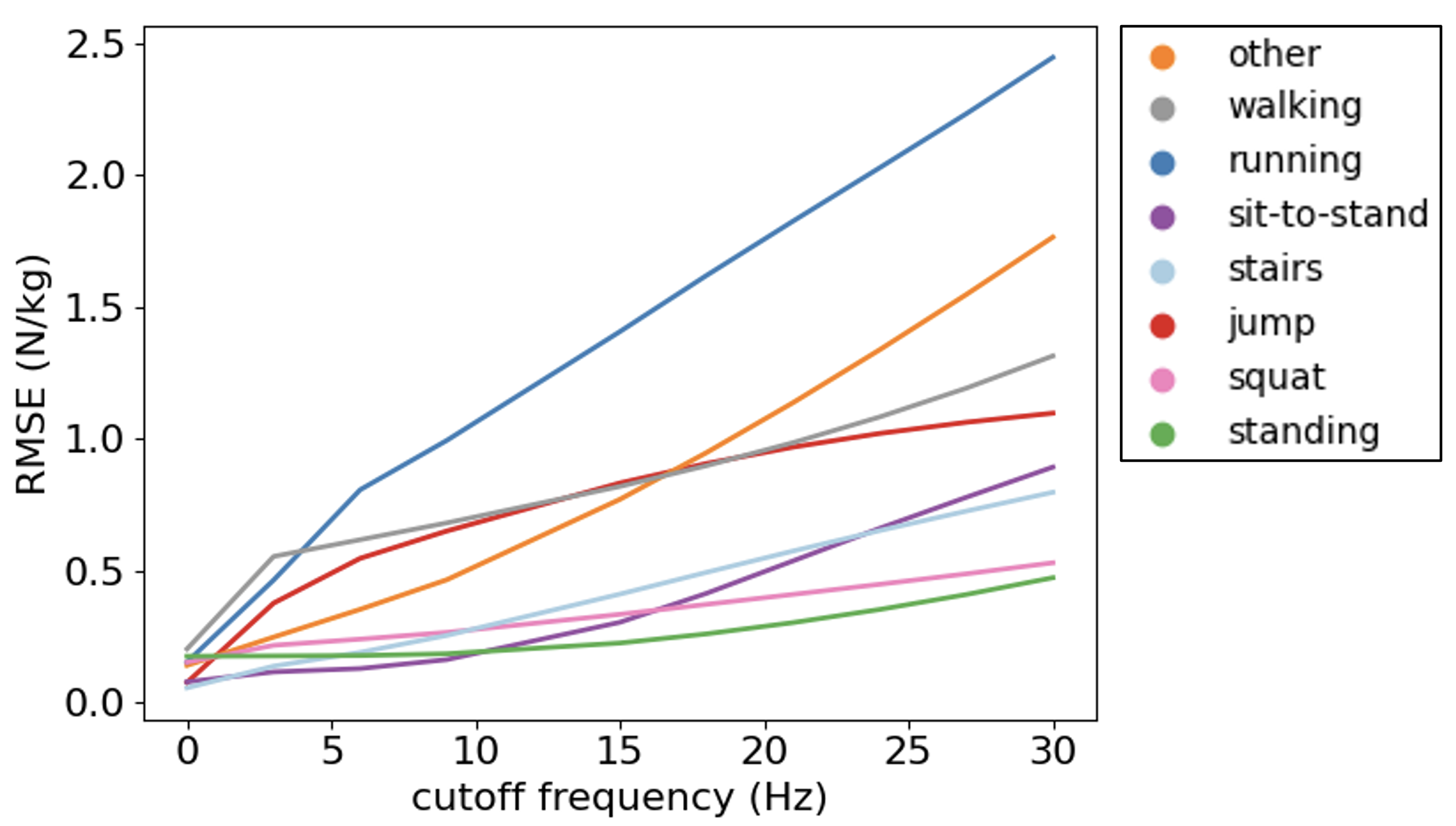}
    \caption{Ground reaction force (GRF) prediction error versus cutoff frequency as shown in Figure \ref{fig:err-v-freq} on the test set, broken down by activity classification.}
    \label{fig:err-v-freq-per-act}
\end{figure}

It is standard practice in biomechanics to low-pass filter experimental data to reduce common sources of noise, such as from electrical noise in the force plates, or measurement noise in motion capture. Figures \ref{fig:err-v-freq} and \ref{fig:err-v-freq-per-act} demonstrates how it is important to use consistent filtering parameters for evaluating downstream ground reaction force (GRF) prediction results. We propose using a 3rd order Butterworth low-pass filter with a 30Hz cutoff frequency as a standard filtering method.


\subsection{AddBiomechanics Dataset Additional Details}
\label{sec:dataset-details}
\begin{figure}
    \centering
    \includegraphics[width=1\linewidth]{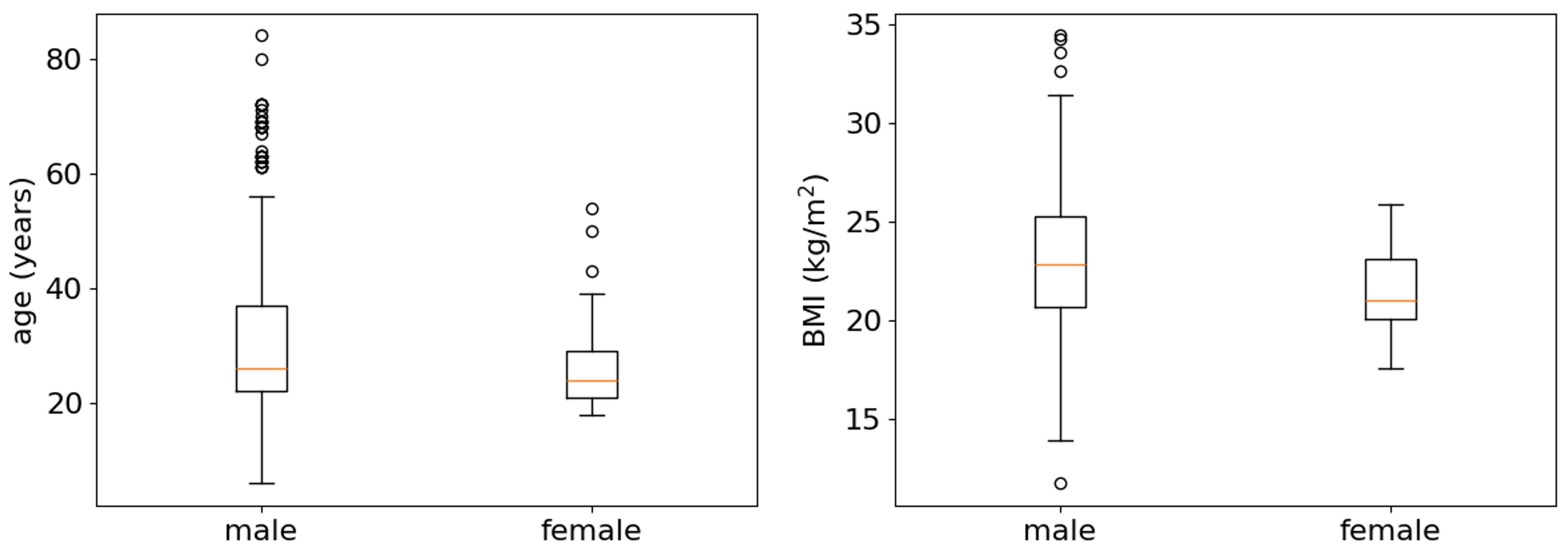}
    \caption{Ages (\textbf{left}) and body mass index (BMI; \textbf{right}) by biological sex in the AddBiomechanics Dataset. Medians are shown by the orange lines. Outliers are shown by the circle markers.}
    \label{fig:demo-boxplots}
\end{figure}

\begin{figure}
    \centering
    \includegraphics[width=0.75\linewidth]{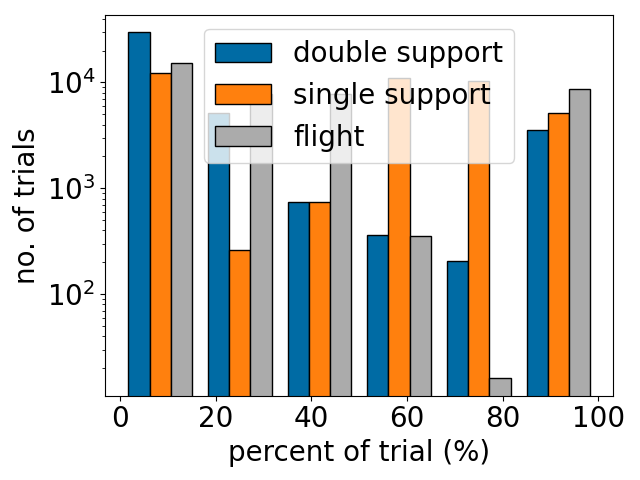}
    \caption{The distribution of the percentage of each trial in the AddBiomechanics Dataset comprised of double support, single support, and flight phase.}
    \label{fig:contact-histo}
\end{figure}

\textbf{Dataset Distributions:} To gain a better understanding of the relative demographic distributions between males and females in the dataset, the breakdown of subject ages and body mass indices (BMI) by biological sex is shown in Figure \ref{fig:demo-boxplots}. Additionally, the distribution of contact phases among trials is shown in Figure \ref{fig:contact-histo}.

\textbf{Activity Classification: }The sub-labels for activity classification include: overground walking, treadmill walking, walking on inclined ramps, overground running, treadmill running, climbing up stairs, descending down stairs, sit-to-stand, jumping, drop jumping, squat jumping, box jumping, squating, lunging, standing, single-support standing, start of transition movements, and end of transition movements. The "other" category includes activities such as yoga positions from \cite{han2023groundlink}. We also classified some trial segments as erroneous.

\textbf{Calculating Average Trial Speeds: }The average speed of each trial was calculated as the average Euclidean norm of the kinematically-derived center of mass (COM) velocity. For treadmill trials however, where the world reference frame is the treadmill, the average speed was calculated by averaging the speed of strides, when either leg was in contact with the treadmill. The speed was calculated as the distance traversed by the ankle joint centers over time in each stride.

\subsection{Dataset Post-Processing Details}
\label{sec:post-processing}
The AddBiomechanics Dataset, output from extensions to Werling \textit{et al.} \cite{werling2023addbiomechanics}, went through additional post-processing steps to prepare the data for machine learning baseline training. The full post-processing pipeline is captured in the open-source code for the Werling \textit{et al.} \cite{werling2023addbiomechanics} tool. Post-processing steps include low-pass filtering, resampling to a standard sampling rate of 100fps, skipping short trials (\textit{e.g.}, less than 12 frames), and using frames with only two contact bodies. 

To ensure accurate dynamics information, we also only use frames that successfully passed through the dynamics optimization of Werling \textit{et al.} \cite{werling2023addbiomechanics}, and only used trials that were manually reviewed by human raters to indicate frames of accurate GRF information (\textit{e.g.}, experimental GRF present when feet are in contact with the ground, separate force vectors in double support, etc.). These criteria helped us ensure that we have a human-verified, accurate and clean data subset for model training, tuning, and evaluation.  For machine learning baseline training, the post-processed dataset was split with an approximately 90:5:5 train:dev:test split, with about 155 subjects represented in the train split, and eight subjects represented each in the dev and test splits. 

The dataset analysis plots we present also use the post-processed data, with further data cleaning to only use frames that were originally flagged as having valid GRF information from automated processing, and excluding trial segments that were identified as erroneous by manual activity classification. Resulting from this further filtering, there are 154 subjects in total, with mean age of 32.7 ($\pm 18.3$) years (range: $6-84$) and mean BMI of 23.1 ($\pm 3.9$) (range: $11.7-34.4$). About 83.8\% of the subjects are male, and 15.6\% of the subjects are female. Age was not reported for four of the subjects, and biological sex was not reported for one of the subjects.

\subsection{Machine Learning Baseline Details}
\label{sec:ml-details}
Table \ref{tab:mlp-hyperparams} summarizes the values of the main hyperparameters from the final MLP baseline model. The final GroundLinkNet used the training-related parameters (\textit{i.e.}, learning rate, optimization algorithm, batch size, and number of epochs) but generally retained the hyperparameters of the original model architecture.
\begin{table}[htbp]
    \centering
    \small
    \begin{tabularx}{\textwidth}{|X|X|X|X|p{1.2cm}|p{1.5cm}|X|p{1cm}|} \hline 
         \textbf{history length}& \textbf{act -ivation}& \textbf{stride length}&  \textbf{hidden nodes}&\textbf{learning rate}&\textbf{opt -imization}&\textbf{batch size}&\textbf{epochs}\\ \hline 
         50&  sigmoid&  5&  512&0.0001 &RMSprop& 32&10\\ \hline
    \end{tabularx}
    \caption{Hyperparameters of final MLP baseline.}
    \label{tab:mlp-hyperparams}
\end{table}

Hyperparameters were manually tuned by mainly assessing different learning rates, optimization algorithms, history lengths, number of epochs, and number of hidden nodes, for a total of 125+ training runs. Training times varied from about five minutes to about 12 hours; the MLP baselines were mainly trained using CPUs locally or on a computing cluster, and the more complicated GroundLinkNet required training with GPUs. Experiments were logged with Weights \& Biases \cite{wandb}.

\subsection{Additional Dataset Analyses}
\label{sec:additional-analyses}
\begin{figure}
    \centering
    \includegraphics[width=1\linewidth]{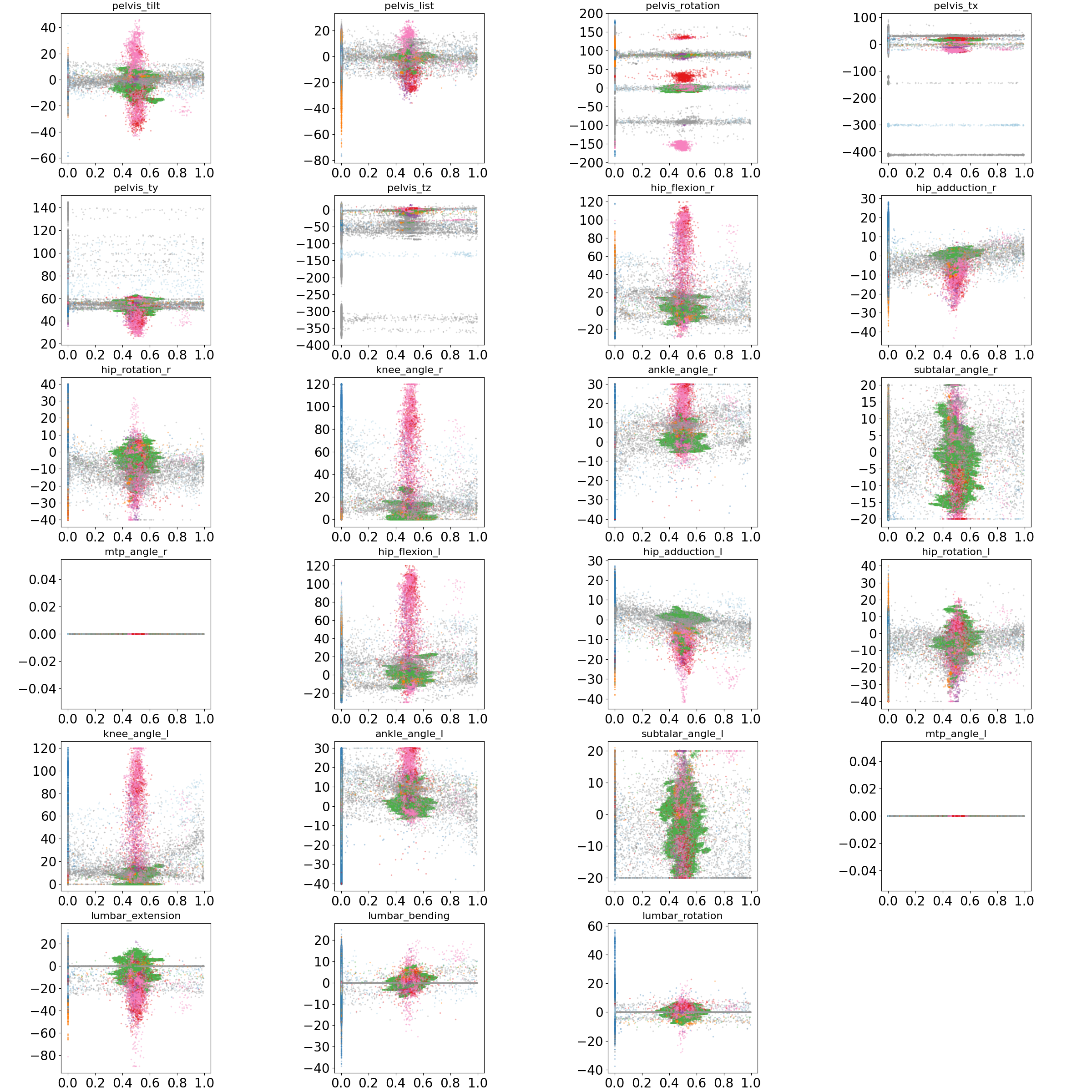}
    \caption{Kinematically-derived joint angles (in degrees) versus the proportion of ground
reaction force (GRF) acting on the right foot. }
    \label{fig:jointpos-dist}
\end{figure}

\begin{figure}
    \centering
    \includegraphics[width=1\linewidth]{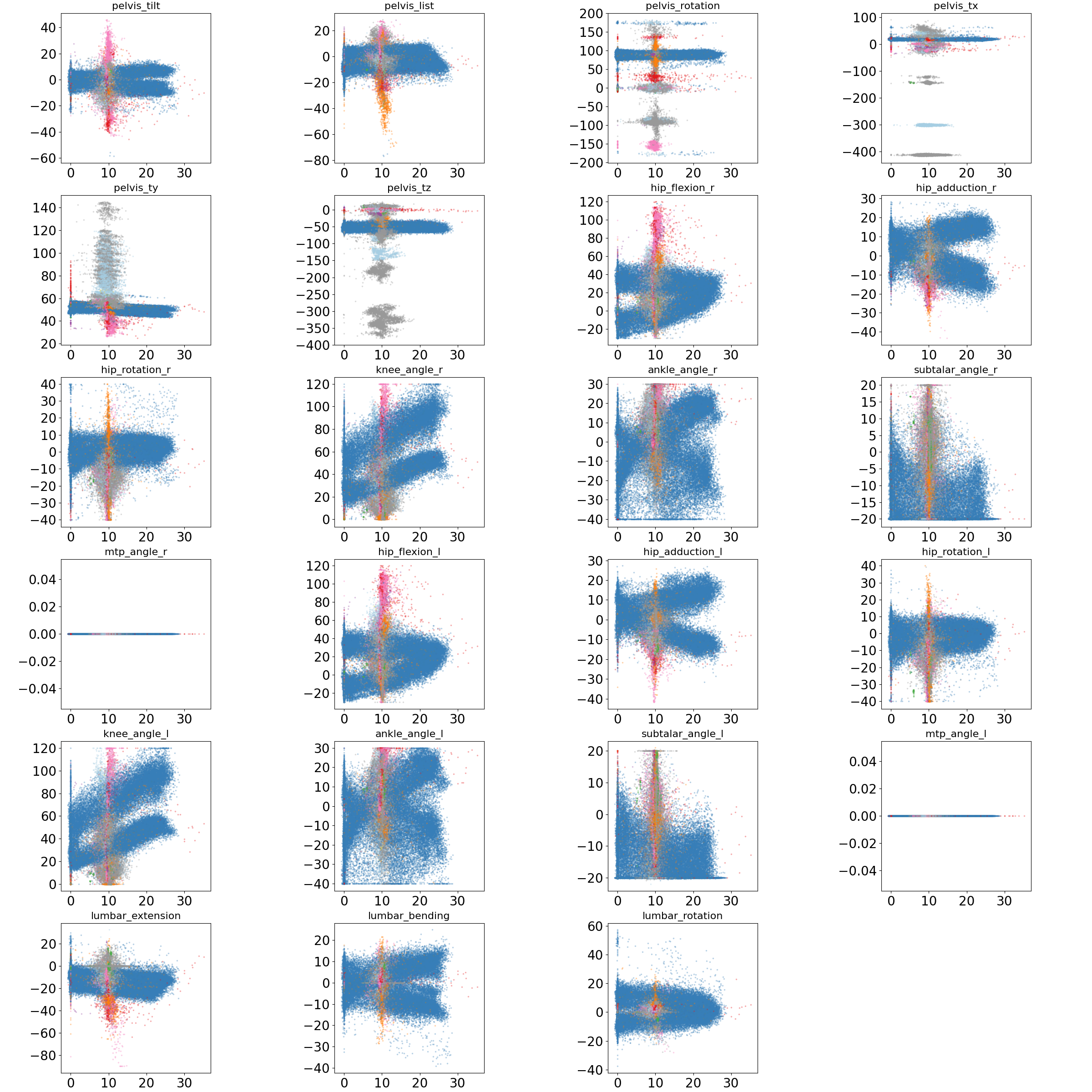}
    \caption{Kinematically-derived joint angles (in degrees) versus total mass-
scaled ground reaction force (GRF) in the vertical direction (in $N/kg$).}
    \label{fig:jointpos-totgrf}
\end{figure}

\begin{figure}
    \centering
    \includegraphics[width=1\linewidth]{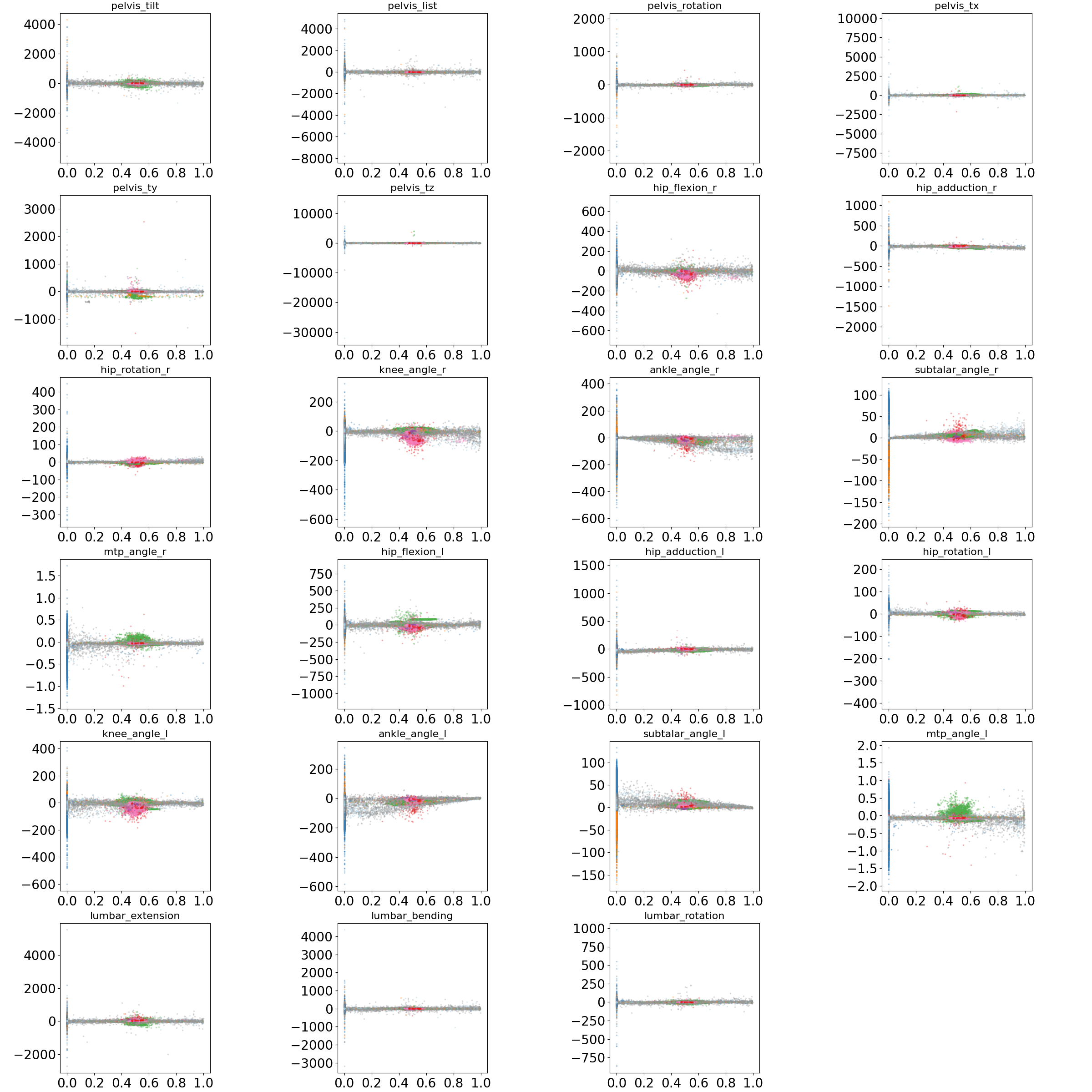}
    \caption{Joint moments (in $Nm$) versus the proportion of ground
reaction force (GRF) acting on the right foot.  }
    \label{fig:jointtau-dist}
\end{figure}

\begin{figure}
    \centering
    \includegraphics[width=1\linewidth]{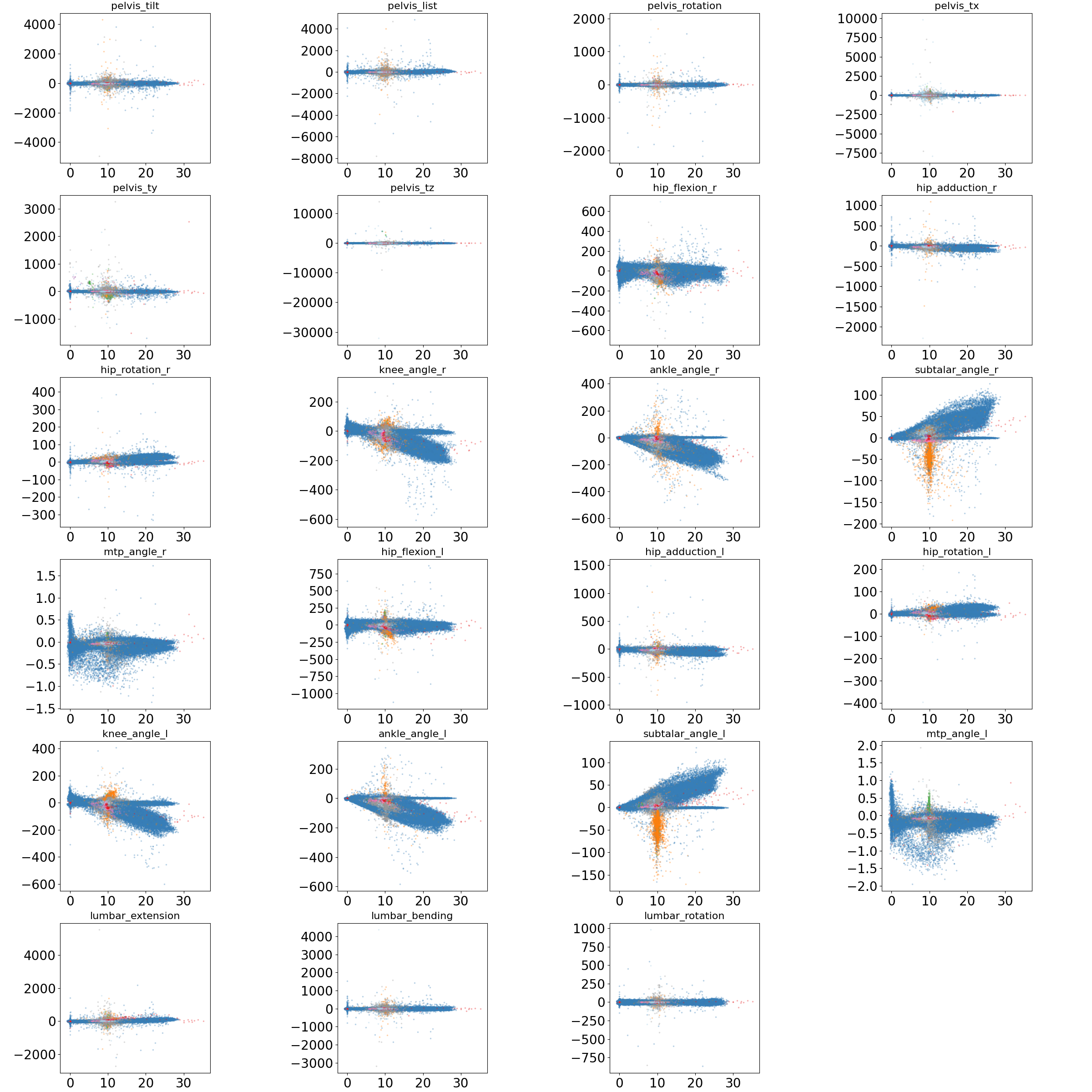}
    \caption{Joint moments (in $Nm$) versus total mass-
scaled ground reaction force (GRF) in the vertical direction (in $N/kg$).}
    \label{fig:jointtau-totgrf}
\end{figure}

\begin{figure}
    \centering
    \includegraphics[width=1\linewidth]{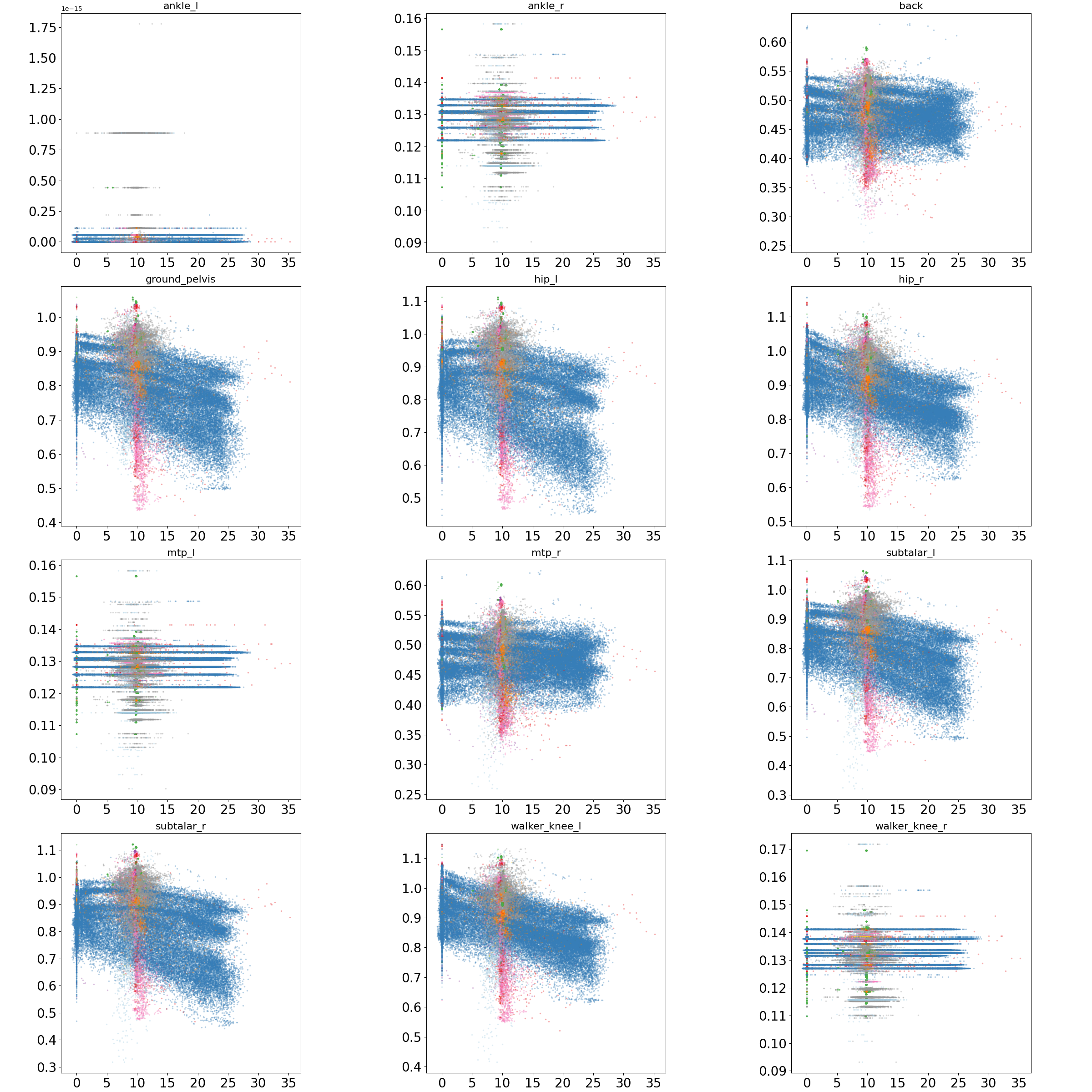}
    \caption{The Euclidean norm of kinematically-derived joint center positions in the pelvis reference frame (in $m$) versus total mass-
scaled ground reaction force (GRF) in the vertical direction (in $N/kg$).}
    \label{fig:jointcenters-totgrf}
\end{figure}

Figures \ref{fig:jointpos-dist}, \ref{fig:jointpos-totgrf}, \ref{fig:jointtau-dist}, \ref{fig:jointtau-totgrf}, and \ref{fig:jointcenters-totgrf} contain comprehensive scatterplots of joint angles, joint moments, or joint center positions versus GRF-related metrics of interest. They use the same activity color scheme as Figure \ref{fig:err-v-freq-per-act} and are down-sampled to 5fps. The degree of freedom and joint center names correspond to the naming conventions used in the provided OpenSim musculoskeletal models \cite{rajagopal2016full}. "\textit{pelvis tx, pelvis ty, pelvis tz}" correspond to the pelvis translational degrees of freedom. "\textit{mtp angle r, mtp angle l}" correspond to the metatarsophalangeal joints, and were locked in the models.

{\small
\bibliographystyle{ieee_fullname}
\bibliography{egbib}
}